\documentclass[10pt,twocolumn,letterpaper]{article}

\usepackage[pagenumbers]{cvpr} 
\usepackage{multirow}
\usepackage{pifont}
\usepackage{algorithm}
\usepackage{algorithmic}
\definecolor{iccvblue}{rgb}{0.21,0.49,0.74}
\usepackage[pagebackref,breaklinks,colorlinks,allcolors=iccvblue]{hyperref}
\usepackage{booktabs}  
\usepackage{amssymb}
\def\paperID{12190} 
\def\confName{CVPR}
\def\confYear{2025}
\begin{document}
\title{UrbanGS: Semantic-Guided Gaussian Splatting for Urban Scene Reconstruction}

\author{Ziwen Li\\
\and
Jiaxin Huang
\and
Runnan Chen
\and
Yunlong Che
\and
Yandong Guo
\and
Tongliang Liu
\and
Fakhri Karray
\and
Mingming Gong
}
\twocolumn[{
\renewcommand\twocolumn[1][]{#1}
\maketitle
\begin{center}
    \captionsetup{type=figure}
    \includegraphics[width=0.99\linewidth]{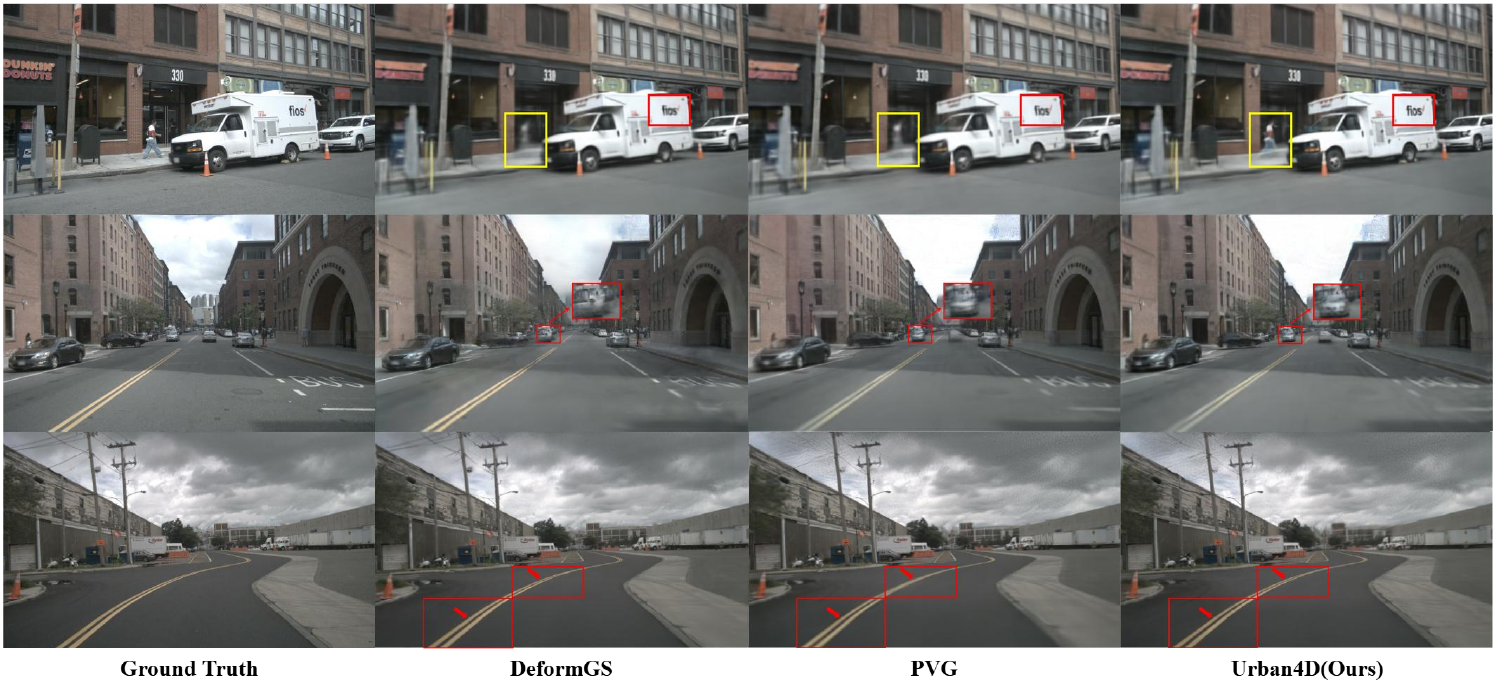}
    \captionof{figure}{Qualitative comparison on the nuScenes~\cite{caesar2020nuscenes} dataset. While DeformGS~\cite{yang2024deformable} achieves comparable results on static regions, it fails on dynamic objects, producing severe artifacts and blurred reconstructions. In contrast, our \textbf{Urban4D} maintains high fidelity for both dynamic objects and static backgrounds, also surpassing the reconstruction quality of PVG~\cite{chen2023periodic}.}
    \label{first}
\end{center}
}]
\begin{abstract}

Reconstructing urban scenes is challenging due to their complex geometries and the presence of potentially dynamic objects. 
3D Gaussian Splatting (3DGS)-based methods have shown strong performance, but existing approaches often incorporate manual 3D annotations to improve dynamic object modeling, which is impractical due to high labeling costs.
Some methods leverage 4D Gaussian Splatting (4DGS) to represent the entire scene, but they treat static and dynamic objects uniformly, leading to unnecessary updates for static elements and ultimately degrading reconstruction quality.
To address these issues, we propose UrbanGS, which leverages 2D semantic maps and an existing dynamic Gaussian approach to distinguish static objects from the scene, enabling separate processing of definite static and potentially dynamic elements.
Specifically, for definite static regions, we enforce global consistency to prevent unintended changes in dynamic Gaussian and introduce a K-nearest neighbor (KNN)-based regularization to improve local coherence on low-textured ground surfaces.
Notably, for potentially dynamic objects, we aggregate temporal information using learnable time embeddings, allowing each Gaussian to model deformations over time.
Extensive experiments on real-world datasets demonstrate that our approach outperforms state-of-the-art methods in reconstruction quality and efficiency, accurately preserving static content while capturing dynamic elements.

\end{abstract}     

\section{Introduction}
\label{sec:intro}

Urban scenes are characterized by two primary categories of objects: major static elements, including buildings and road infrastructure, which remain spatially consistent over time, and some potentially dynamic elements, such as pedestrians and vehicles, which can remain static or exhibit diverse and often unpredictable motion patterns. Accurate reconstruction of urban scenes thus remains challenging, mainly due to the coexistence of these static and dynamic elements and complexities arising from low-textured regions.

Recent advancements in 3D Gaussian Splatting (3DGS) ~\cite{yan2024street,zhou2024drivinggaussian,chen2024omnire,fischer2024dynamic}, which have attempted to incorporate manually labeled 3D bounding boxes to process dynamic objects separately. However, such manual annotations are labor-intensive, impractical for large-scale settings, and unsuitable for continuously evolving environments where dynamic objects frequently change positions. 
Alternative approaches leverage 4DGS-based representations, such as Periodic Vibration Gaussian (PVG)~\cite{chen2023periodic}, which introduces periodic temporal modeling to represent motion variations in urban scenes. However, these methods lack explicit differentiation between static and dynamic elements, leading to unnecessary updates for stationary objects, which in turn degrades reconstruction quality.

We observe that Gaussians supervised by semantic maps inherently acquire semantic information, which can be leveraged for identifying static objects. Furthermore, deep-learning-based 2D semantic segmentation models provide robust classification capabilities, allowing Gaussians to be categorized into determined static and potentially dynamic elements without relying on explicit 3D annotations.

Inspired by this perspective, we introduce \textbf{UrbanGS}, a semantic-guided Gaussian Splatting framework designed to effectively handle definite static elements in urban scene reconstruction while adapting to potentially dynamic components. Specifically, for definite static Gaussians, we introduce a global consistency constraint to ensure that they remain unchanged over time. Additionally, we employ a K-nearest neighbor (KNN)-based consistency regularization to improve local coherence, particularly for low-textured surfaces such as roads, which pose significant challenges in urban scene reconstruction.
For potentially dynamic Gaussians, we propose an efficient 4DGS representation that incorporates learnable time embeddings for each Gaussian. This design enables the model to predict object deformations at arbitrary timestamps using a lightweight multilayer perceptron (MLP), effectively capturing urban dynamics while preserving rendering efficiency.


Our extensive experiments on real-world urban datasets demonstrate that UrbanGS achieves state-of-the-art reconstruction quality in both static and potentially dynamic objects. The key contributions of this work can be summarized as follows:
\begin{enumerate}
\item[$\bullet$] We introduce UrbanGS, a novel semantic-driven framework that leverage 2D semantic segmentation to separate static Gaussians from potentially dynamic Gaussians without requiring manual 3D annotations.

\item[$\bullet$] 
We propose a global consistency constraint to enforce temporal stability in static Gaussians, preventing unnecessary updates and significantly improving reconstruction quality. Additionally, to address the challenge of low-textured regions, we introduce a KNN-based consistency regularization, ensuring a more stable and accurate reconstruction of surfaces such as roads and sidewalks.

\item[$\bullet$] 
We develop a learnable time embedding mechanism for potentially dynamic Gaussians, enabling the model to predict object deformations at arbitrary timestamps using a lightweight MLP-based deformation model, efficiently handling motion in urban scenes.
\end{enumerate}
\begin{figure*}[!t]
    \centering
    \includegraphics[width=0.90\linewidth]{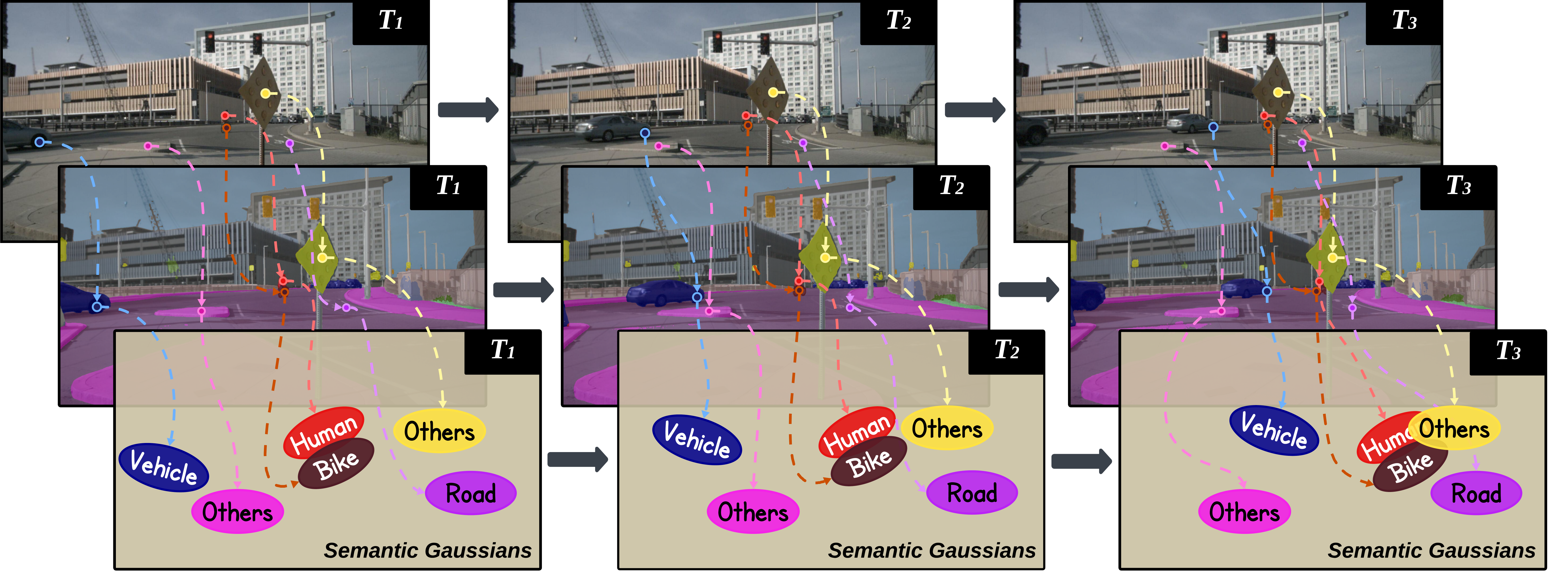}
    \caption{Semantic-guided decomposition over time. For each timestamp ($T_1$, $T_2$, $T_3$), semantic Gaussians of the current frame are obtained through rendering and supervision of corresponding semantic maps. Dynamic classes include vehicles, pedestrians, and cyclists, while the static set comprises buildings, vegetation, and roads. For simplicity, we use the "Road" to represent ground surfaces.}
    \label{fig:third}
    \vspace{-3mm}
\end{figure*}
\section{Related Work}

\noindent \textbf{Neural Scene representations} have revolutionized novel view synthesis, with NeRF~\cite{mildenhall2021nerf} leading significant advances in this field. NeRF utilizes multi-layer perceptrons (MLPs) and differentiable volume rendering to reconstruct 3D scenes from 2D images and camera poses. While demonstrating impressive results for bounded scenes, its application to large-scale unbounded scenes remains challenging due to computational constraints and the requirement for consistent camera-object distances.

Various improvements have been proposed to address NeRF's limitations. Training speed has been enhanced through techniques like voxel grids~\cite{garbin2021fastnerf,sun2022direct,fridovich2022plenoxels}, hash encoding~\cite{mueller2022instant}, and tensor factorization~\cite{chen2022tensorf}, while rendering quality has been improved through better anti-aliasing~\cite{barron2023zip,kopanas2022neural,liu2024rip} and reflection modeling~\cite{verbin2022ref,guo2022nerfren}.

More recently, 3D Gaussian Splatting~\cite{kerbl20233d} has emerged as a promising alternative, offering faster training and rendering while maintaining high quality results. This explicit representation combines the advantages of volumetric rendering with efficient rasterization-based techniques. Compared to previous explicit representations (\eg, mesh, voxels), 3D-GS can model complex shapes while allowing fast, differentiable rendering through splat-based rasterization. 

\noindent \textbf{Dynamic scene reconstruction} methods generally fall into two categories: deformation-based and modulation-based approaches. Deformation-based methods~\cite{park2021nerfies,park2021hypernerf,cai2022neural,tretschk2021non,pumarola2021d} model scene dynamics through canonical space mapping and deformation networks, while modulation-based approaches~\cite{li2021neural,li2022neural,xian2021space,luiten2023dynamic} incorporate temporal information directly. These methods have shown promising results in controlled environments but face significant challenges when applied to complex real-world scenarios with multiple dynamic objects.

For urban environments, several pioneering works have tackled static scene reconstruction by introducing multi-scale NeRF variants~\cite{martin2021nerf,tancik2022block,turki2022mega} and incorporating advanced rendering techniques like mipmapping~\cite{barron2021mip,barron2022mip}. Building upon these foundations, recent methods~\cite{xie2023s,turki2023suds,yang2023emernerf} have explored the integration of multi-modal data, combining RGB images with LiDAR point clouds to enhance geometric accuracy. However, the challenge of jointly modeling static and dynamic elements remains complex, particularly due to high-speed movements and sparse viewpoints typical in driving scenarios.

To address these challenges, recent works have proposed various scene decomposition strategies. Scene graph representations~\cite{yang2023unisim,wu2023mars,yan2024street,zhou2024drivinggaussian,tonderski2024neurad,chen2024omnire,fischer2024dynamic} enable explicit modeling and control at the object level. However, most current approaches either treat all dynamic elements uniformly~\cite{chen2023periodic,yang2024deformable,huang2024s3gaussian} or rely heavily on manual annotations~\cite{zhou2024drivinggaussian,yan2024street,chen2024omnire,fischer2024dynamic}.

\noindent \textbf{Gaussian reconstruction with semantic features}
Recent advances integrate 3DGS with semantic features. Feature 3DGS~\cite{zhou2024feature} extends 3D Gaussian Splatting by introducing high-dimensional feature fields. Similarly, Semantic Gaussians~\cite{guo2024semantic} tackles open-vocabulary 3D scene understanding by mapping diverse 2D semantic features into 3D Gaussian. These works focus on static scenes, while our work complements these efforts by focusing on urban scenes, leveraging semantic decomposition for static/dynamic separation.

\section{Methodology}

\begin{figure*}[!ht]
\begin{minipage}[b]{1.0\linewidth}
  \centering
  \centerline{\includegraphics[width=17.5cm]{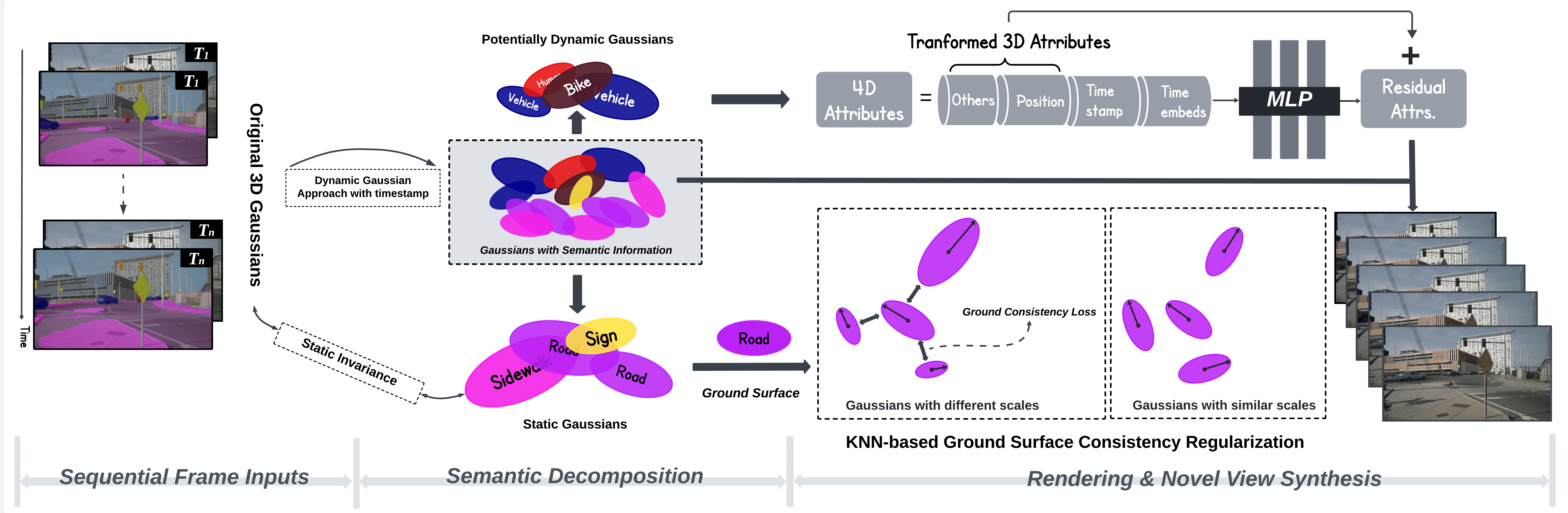}}
\end{minipage}
\caption{Overview of \textbf{UrbanGS} framework. Given input images with semantic information during training, Gaussians are classified into definite static and potentially dynamic elements through semantic-guided decomposition. For definitively static Gaussians, we introduce a static invariance constraint to preserve their temporal invariance and prevent unintended transformations. To address challenges in low-texture regions (e.g., ground surfaces), a KNN-based regularization mechanism is employed to enforce structural coherence. Potentially dynamic objects are represented in 4D Gaussian Splatting that captures motion patterns by incorporating a learnable time embedding, with deformations predicted at desired timestamps using an MLP.}
\label{Main}
\end{figure*}

\noindent\textbf{Method overview.} Our proposed method aims to reconstruct dynamic urban scenes by leveraging semantic information to effectively distinguish between determined static and potentially dynamic elements. 

Given a sequence of images $\{I_t\}_{t=1}^T$ and the corresponding LiDAR point clouds $\{P_t\}_{t=1}^T$ captured by a moving vehicle, our aim is to reconstruct urban scenes. For each frame, semantic maps $\{S_t\}_{t=1}^T$ are predicted using an off-the-shelf pre-trained segmentation model. Building upon a uniformly dynamic Gaussian approach, which inherently applies time-dependent transformations to the global Gaussians, such as PVG~\cite{chen2023periodic}, we propose a novel framework that introduces semantic-aware improvement. Specifically, during training, we leverage the semantic attributes of each Gaussian to enforce constraints and adaptively adjust the properties of each Gaussian point. This allows us to effectively keep static elements unchanged across time, enforce consistency in low-texture regions, and intuitively capture potentially dynamic objects through a 4D representation. Our approach enhances the robustness and accuracy of urban scene reconstruction by combining the strengths of semantic guidance and dynamic Gaussian modeling. More details of pseudo algorithm for training are included in the supplementary material.

Our method consists of three main elements (Figure~\ref{Main}): (1) semantic-guided decomposition that separates the scene into static and potentially dynamic Gaussians based on semantic information (Sec.~\ref{sec:decomposition}), (2) a temporal-invariance regularization for all static points, ensuring they remain unchanged over time, and apply a KNN-based consistency constraint to low-texture regions for enhanced reconstruction fidelity (Sec.~\ref{sec:static}), and (3) a 4d Gaussians Splatting representation for potentially dynamic objects (Sec.~\ref{sec:dynamic}) .

\subsection{Preliminaries}



3D Gaussian Splatting represents a scene as a set of 3D 
Gaussians $\{\mathcal{G}_i\}_{i=1}^N$, where each Gaussian 
$\mathcal{G}_i$ is parameterized by its mean position 
$\boldsymbol{\mu}_i \in \mathbb{R}^3$, covariance matrix 
$\boldsymbol{\Sigma}_i \in \mathbb{R}^{3\times3}$, and 
appearance features including opacity $\alpha_i \in 
\mathbb{R}$ and spherical harmonics coefficients $\mathbf{f}_i 
\in \mathbb{R}^{48}$ for RGB color representation.

For each pixel in the target view, the rendering process involves projecting the 3D Gaussians onto the 2D image plane. The projection of a 3D Gaussian results in a 2D Gaussian with parameters:
\begin{equation}
    \boldsymbol{\mu}_{2D} = \Pi(\boldsymbol{\mu}_i),
\end{equation}
\begin{equation}
    \boldsymbol{\Sigma}_{2D} = J\boldsymbol{\Sigma}_iJ^T,
\end{equation}
where $\Pi(\cdot)$ is the perspective projection function and $J$ is the Jacobian of the projection.

The final color $C(x,y)$ at pixel $(x,y)$ is computed through alpha compositing:
\begin{equation}
    C(x,y) = \sum_{i=1}^N T_i\alpha_i c_i,
\end{equation}
where $T_i$ represents the accumulated transmittance, $\alpha_i$ is the opacity, and $c_i$ is the RGB color from spherical harmonics coefficients. The scene is optimized by minimizing the difference between rendered and ground-truth images.

In addition to RGB supervision, each Gaussian can be associated with semantic labels $s_i \in \{1,...,K\}$, where $K$ is the number of semantic classes. The semantic prediction at pixel $(x,y)$ can be computed similarly through alpha compositing:
\begin{equation}
    S(x,y) = \sum_{i=1}^N T_i\alpha_i s_i,
\end{equation}

\subsection{Semantic-guided Decomposition}
\label{sec:decomposition}

Leveraging 2D semantic maps from a pre-trained segmentation model, we introduce a systematic approach to identify and separate static elements, allowing for specialized treatment of static elements and remaining potentially movable objects. Specifically, as shown in Figure~\ref{fig:third}, during dynamic Gaussian training, each Gaussian point $\mathcal{G}_i$ is assigned a semantic label $s_i \in \{1,...,K\}$, obtained through rendering and supervision of semantic maps, where $K$ represents the total number of semantic classes.
Our method focuses on determined static objects in urban scenes, typically comprising well-defined semantic categories such as building, roads and trees. We have designed a strategy specifically for those static objects to ensure their invariance over time. We also identify potentially dynamic classes $\mathcal{C}_d$, including vehicles, pedestrians, and cyclists, which frequently exhibit motion and require specialized handling.
This semantic understanding enables the decomposition of the scene into two disjoint sets:
\begin{equation}
\begin{aligned}
    \mathcal{G}_i^d = \{\mathcal{G}_i | s_i \in \mathcal{C}_d\}, \\
    \mathcal{G}_i^s = \{\mathcal{G}_i | s_i \in \mathcal{C}_s\},
\end{aligned}
\end{equation}
where $\mathcal{G}_i^s$ denotes the static Gaussians that require static regularization, while $\mathcal{G}_i^d$ refers to dynamic Gaussians (i.e., Gaussians associated with potentially dynamic objects) that necessitate 4D modeling.
This semantic-guided decomposition offers several key advantages: (1) it ensures static elements remain unchanged over time while confining temporal modeling to potentially dynamic objects, (2) it enhances the reconstruction quality specifically for road, and (3) it eliminates the need for labor-intensive manual annotations.

The decomposition lays the foundation for our two-stream optimization strategy. the static Gaussians receive geometric regularization (Sec.~\ref{sec:static}) to enhance scene stability, while the potentially dynamic Gaussians in $\mathcal{G}_i^d$ undergo a dedicated motion refinement (Sec.~\ref{sec:dynamic}) to accurately capture movement. Additionally, we employ an optimizable environment texture map for sky representation, which is rendered separately and combined with the Gaussian-based image by alpha blending, as described in~\cite{chen2023periodic}. 

\subsection{Static Regularization}
\label{sec:static}
\noindent\textbf{Static invariance.} Prior approaches (e.g., PVG) address dynamic scene reconstruction by applying timestamp-dependent transformations to each Gaussian’s 3D position $\mu$ and opacity $\alpha$. These transformations effectively capture dynamic motions but inevitably alter truly \textbf{static}
parts of the scene. To mitigate this issue, we introduce a static consistency loss to keep these static Gaussians invariant:
\[
\mathcal{L}_\text{static}
= \sum_{i \in \mathcal{G}_i^s}
w_i \Bigl(\|\mu_i - \mu_i'\|^2 
+ \|\alpha_i - \alpha_i^\prime\|^2\Bigr),
\]
where  $\mu_i$ and $\alpha_i$ denote the untransformed parts of static Gaussians, and ${\mu_i'}$ and $\alpha_i'$ are their transformed counterparts. \(w_i\) is a semantic weight (derived from a softmax ratio) indicating 
the likelihood that Gaussian \(i\) is truly static. This weighting mechanism 
allows fully static points (\(w_i \approx 1\)) to remain nearly unchanged, 
while points that are partly dynamic or uncertain (\(0 < w_i < 1\)) retain 
the freedom to move if necessary.

\noindent\textbf{Ground surface consistency regularization.} In urban driving scenes, ground surfaces constitute a significant portion of the environment and typically exhibit low-texture characteristics. While ground-level Gaussians should theoretically share similar properties due to their homogeneous nature, enforcing strict uniformity across all ground Gaussians would be oversimplified and impractical, as real-world surfaces often contain variations and irregularities. 
The scale parameter of a Gaussian, derived from its covariance matrix, inherently encodes local geometric information analogous to surface normals~\cite{jiang2024gaussianshader,cheng2024gaussianpro}. A well-behaved ground surface should exhibit smooth transitions in its local geometry, making scale a particularly suitable target for regularization. This motivates us to regularize the scale parameters rather than other Gaussian properties. 

For each ground Gaussian $\mathcal{G}_i \in \mathcal{G}_g$ (where $\mathcal{G}_g \subset \mathcal{G}_i^s$ denotes the set of ground surface Gaussians), we identify its $N$ nearest neighbors:
\begin{equation}
    \mathcal{N}_i = \text{KNN}(\mathcal{G}_i, \mathcal{G}_g, N),
\end{equation}
where KNN retrieves the $N$ spatially closest Gaussians to $\mathcal{G}_i$ from the global set $\mathcal{G}_g$. The neighbors are determined based on the Euclidean distance between Gaussian centers $\boldsymbol{\mu}_i$, forming a local neighborhood for geometric consistency.
We then introduce a local consistency loss that encourages similar scale properties within each local neighborhood:
\begin{equation}
\label{eq:ground_loss}
    \mathcal{L}_{\text{ground}} = \sum_{\mathcal{G}_i \in \mathcal{G}_g} \|\boldsymbol{s}_i - \frac{1}{N}\sum_{\mathcal{G}_j \in \mathcal{N}_i} \boldsymbol{s}_j\|_2^2,
\end{equation}
where $\boldsymbol{s}_i$ and $\boldsymbol{s}_j$ represents the scale parameter of the Gaussian $\mathcal{G}_i$ and $\mathcal{G}_j$. By regularizing the scale parameters, we effectively enforce consistency in the local surface geometry while preserving the ability to model natural surface variations. This approach leads to more coherent ground surface reconstruction, as similar scale parameters in a local neighborhood implicitly enforce consistent surface normal orientations, resulting in improved geometric fidelity of the ground surface representation.

\subsection{4D Gaussian Splatting Representation}
\label{sec:dynamic}
While the original dynamic Gaussian approach demonstrates the capability to model dynamic objects, we aim to refine it further to better handle the remaining potentially dynamic Gaussians. To refine these potentially dynamic Gaussians ${\mathcal{G}_i^d}$ in 4D, our method extends the deformation mechanism of DeformGS~\cite{yang2024deformable} by introducing a learnable time embedding for each Gaussian. Unlike DeformGS~\cite{yang2024deformable}, which directly maps spatial positions and time to deformations, our approach leverages temporal context through Gaussian-specific embeddings. This refinement enables a more adaptive representation of dynamic elements.

Specifically, for each dynamic Gaussian $\mathcal{G}_i^d$, we maintain a learnable time embedding vector $\mathbf{e}_i \in \mathbb{R}^{D_e}$. At time step $t$, we form the input feature by concatenating this temporal embedding with position and time information:
\begin{equation}
    \mathbf{h}_i(t) = [\boldsymbol{\mu}_i; t; \mathbf{e}_i],
\end{equation}
where $[;]$ denotes concatenation, $t$ is the normalized time stamp, and $\boldsymbol{\mu}_i \in \mathbb{R}^3$ represents the original 3D position of the Gaussian. This temporal-aware design enables more accurate modeling of complex motions compared to the direct mapping used in DeformGS~\cite{yang2024deformable}.

This combined feature vector is processed by a lightweight MLP to predict residual corrections:
\begin{equation}
    [\Delta\boldsymbol{\mu}_i(t),  \Delta\alpha_i(t),\Delta\boldsymbol{r}_i(t),\Delta\boldsymbol{s}_i(t)] = \text{MLP}(\mathbf{h}_i(t)),
\end{equation}
where $\Delta\boldsymbol{\mu}_i(t) \in \mathbb{R}^3$, $\Delta\alpha_i(t) \in \mathbb{R}$, $\Delta\boldsymbol{r}_i(t) \in \mathbb{R}^4$, $\Delta\boldsymbol{s}_i(t) \in \mathbb{R}^3$ are the predicted position, opacity, rotation and scale residuals, respectively. The final parameters of 4D Gaussians at time $t$ are obtained by:
\begin{equation}
\begin{aligned}
    \boldsymbol{\mu}_i'(t) &= \boldsymbol{\mu}_i + \Delta\boldsymbol{\mu}_i(t), \\
    \alpha_i'(t) &= \alpha_i + \Delta\alpha_i(t), \\
    \boldsymbol{r}_i'(t) &= \boldsymbol{r}_i + \Delta\boldsymbol{r}_i(t), \\
    \boldsymbol{s}_i'(t) &= \boldsymbol{s}_i + \Delta\boldsymbol{s}_i(t), \\ 
\end{aligned}
\end{equation}
where $\boldsymbol{\mu}_i \in \mathbb{R}^3$, $\alpha_i \in [0,1]$, $\boldsymbol{r}_i \in \mathbb{R}^3$,$\boldsymbol{s}_i \in \mathbb{R}^3$ denotes the initial 3D position, opacity, rotation and scaling of the $i$-th Gaussian.
These refined parameters are then used in the standard 3D Gaussian Splatting rendering process to generate the final images. This refinement mechanism allows each dynamic Gaussians to adapt its attributions based on its temporal context, enabling more accurate representation of moving objects in the scene.

The MLP architecture is intentionally kept lightweight to maintain computational efficiency while providing sufficient capacity for modeling temporal dynamics. The detailed architecture of MLPs used in our method is provided in the supplementary material for reproducibility. The entire refinement process is end-to-end trainable along with the main 3DGS optimization objectives.

\subsection{Optimization Strategy}
\label{sec:optimization}

Our optimization objective comprises multiple loss terms that jointly ensure high-quality visual rendering, geometric accuracy, and semantic consistency. The overall loss function is formulated as follows:
\begin{align}
    \mathcal{L} = & \lambda_1\mathcal{L}_{\text{L1}}+ \lambda_2\mathcal{L}_{\text{SSIM}}+\lambda_3\mathcal{L}_{\text{sem}} \nonumber \\
    & + \lambda_4\mathcal{L}_{\text{static}} + \lambda_5\mathcal{L}_{\text{ground}} + \lambda_6\mathcal{L}_{\text{depth}} + \lambda_7\mathcal{L}_{\text{sky}},
\end{align}
where $\{\lambda_i\}_{i=1}^7$ are weighting coefficients balancing different loss terms. Each loss term serves a specific purpose in our optimization.

\noindent\textbf{Appearance Supervision.}
The L1 loss and SSIM loss work together to ensure accurate color reproduction and structural similarity:
\begin{equation}
\begin{aligned}
    \mathcal{L}_{\text{L1}} &= \|I_{\text{rendered}} - I_{\text{gt}}\|_1, \\
    \mathcal{L}_{\text{SSIM}} &= 1 - \text{SSIM}(I_{\text{rendered}}, I_{\text{gt}}), \\ 
\end{aligned}
\end{equation}
where $I_{\text{rendered}}$ and $I_{\text{gt}}$ denote the rendered image and ground-truth image respectively.

\noindent\textbf{Semantic Consistency.}
The semantic loss ensures correct class predictions for each Gaussian:
\begin{equation}
    \mathcal{L}_{\text{sem}} = \text{CE}(S_{\text{rendered}}, S_{\text{gt}}),
\end{equation}
where CE denotes cross-entropy loss between rendered semantic maps $S_{\text{rendered}}$ and ground-truth semantic maps $S_{\text{gt}}$.

\noindent\textbf{Static invariance \& Ground Consistency.}
As detailed in Sec.~\ref{sec:static}.

\noindent\textbf{Geometric Supervision.}
The inverse depth loss aligns the scene geometry with LiDAR measurements:
\begin{equation}
    \mathcal{L}_{\text{depth}} = \|\frac{1}{D_{\text{rendered}}} - \frac{1}{D_{\text{lidar}}}\|_1,
\end{equation}
where $D_{\text{rendered}}$ and $D_{\text{lidar}}$ represent the rendered depth and LiDAR depth respectively.

\noindent\textbf{Sky Region Handling.}
For sky regions, we encourage low opacity to prevent incorrect geometry:
\begin{equation}
    \mathcal{L}_{\text{sky}} = \sum_{\mathcal{G}_i \in \mathcal{G}_{\text{sky}}} \|\alpha_i\|_1,
\end{equation}
where $\mathcal{G}_{\text{sky}}$ represents the set of sky Gaussians.

These joint loss terms collectively constrain the scene reconstruction process, ensuring high-quality results.

\begin{figure*}[!t]
    \centering
    \includegraphics[width=\linewidth]{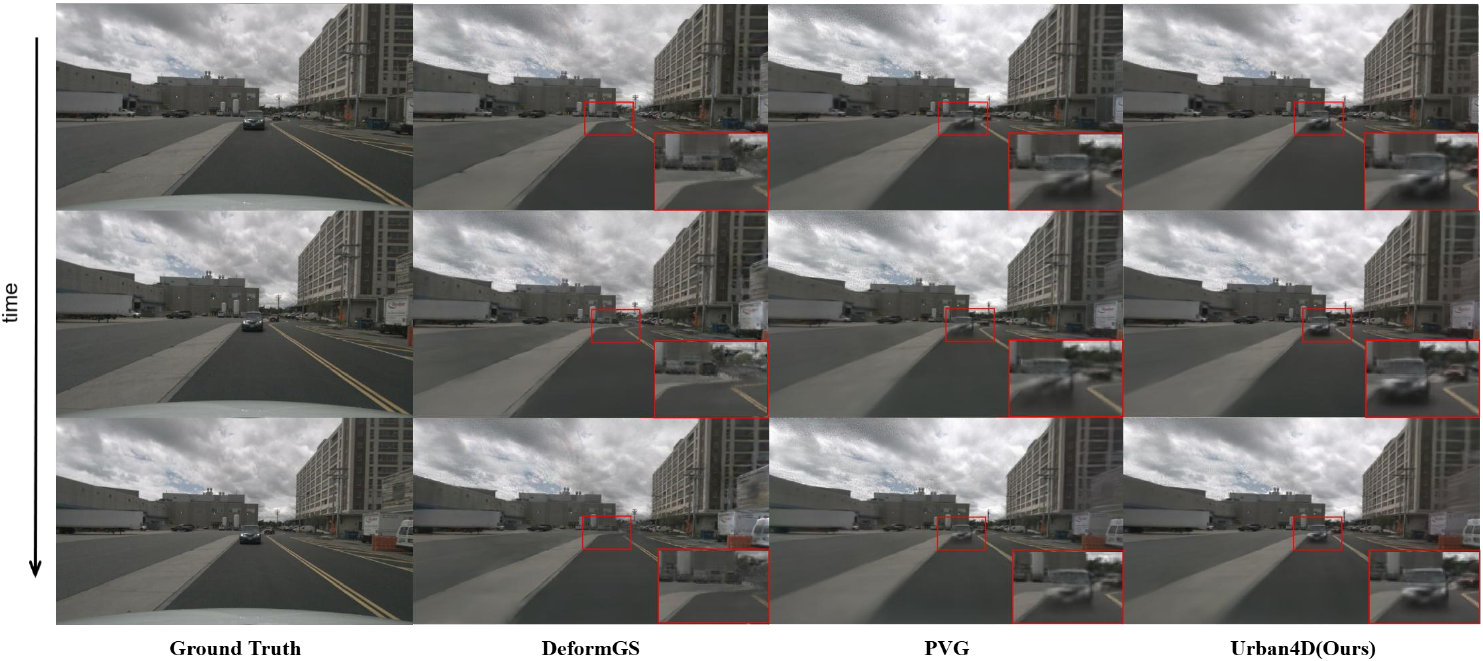}
    \caption{Comparison of reconstruction quality across consecutive frames. DeformGS~\cite{yang2024deformable} struggles significantly with reconstructing dynamic objects, resulting in severe artifacts and a failure to accurately represent motion. PVG~\cite{chen2023periodic} captures dynamic vehicles to some extent but suffers from noticeable blurring, particularly in the lower parts of the objects. In contrast, \textbf{UrbanGS} delivers superior reconstruction quality, maintaining high fidelity and preserving clear details throughout the dynamic objects.}
    \label{fig:comparison}
\end{figure*}

\begin{table*}[!t]
    \fontsize{9}{10}\selectfont
    \centering
    \renewcommand\arraystretch{1.2}
    \caption{Quantitative comparison with existing methods on the nuScenes~\cite{caesar2020nuscenes} dataset. \textbf{D}: DINO features, \textbf{F}: Optical Flow, \textbf{S}: Semantic map, \textbf{B}: Bounding box. * indicates methods that use ground-truth (GT) bounding boxes.}
    \label{tab:comparison}
    \small
    \resizebox{\textwidth}{!}{%
    \begin{tabular}{l|c|c|c|cc|cc|cc|cc}
    \toprule
    \multirow{2}{*}{Method} & \multirow{2}{*}{Extra Inputs} & \multicolumn{2}{c|}{Full Image} & \multicolumn{2}{c|}{Non-Sky} & \multicolumn{2}{c|}{Human} & \multicolumn{2}{c|}{Vehicle} & \multicolumn{2}{c}{Dynamic} \\
    & & PSNR$\uparrow$ & SSIM$\uparrow$ & PSNR$\uparrow$ & SSIM$\uparrow$ & PSNR$\uparrow$ & SSIM$\uparrow$ & PSNR$\uparrow$ & SSIM$\uparrow$ & PSNR$\uparrow$ & SSIM$\uparrow$ \\
    \midrule
    SUDS~\cite{turki2023suds}& D+F & 20.02 & 0.605 & 20.45 & 0.621 & 21.20 & 0.633 & 21.98 & 0.678 & 21.49 & 0.646 \\
    DeformGS~\cite{yang2024deformable} &  & 25.32 & 0.825 & 25.27 & 0.822 & 21.44 & 0.6456 & 21.72 & 0.700 & 21.62 & 0.684 \\
    3DGS~\cite{kerbl20233d} & & 26.02 & 0.825 & 26.45 & 0.836 & 23.20 & 0.721 & 23.98 & 0.794 & 23.59 & 0.756 \\
    EmerNeRF~\cite{yang2023emernerf} & S &  26.12 & 0.830 & 26.50 & 0.840 & 23.45 & 0.733 & 24.69 & 0.808 & 24.12 & 0.767 \\
    PVG~\cite{chen2023periodic} & S & 26.23 & 0.834 & 26.72 & 0.841 & 23.98 & 0.743 & 24.73 & 0.815 & 24.33 & 0.774 \\
    streetGS~\cite{yan2024street} & S+B & 25.46 & 0.831 & 25.50 & 0.820 & 22.54 & 0.659 & 25.87 & 0.837 & 24.17 & 0.771 \\
    4dgf~\cite{fischer2024dynamic} & B & 21.20 & 0.694 & 24.37 & 0.769 & 20.36 & 0.619 & 22.74 & 0.762 & 22.39 & 0.698 \\
    OmniRe~\cite{chen2024omnire} & S+B & 26.41 & 0.837 & 26.73 & 0.845 & 23.71 & 0.737 & 25.95 & \textbf{0.856} & 25.24 & 0.805 \\
    \textbf{UrbanGS} (Ours) & S & \textbf{26.92} & \textbf{0.848} & \textbf{27.61} & \textbf{0.861} & \textbf{24.92} & \textbf{0.767} & \textbf{26.01} & 0.838 & \textbf{25.43} & \textbf{0.818} \\
    \midrule
    streetGS~\cite{yan2024street} & S+$\mathrm{B}^*$ & 25.89 & 0.845 & 26.01 & 0.858 & 22.83 & 0.705 & 26.88 & 0.852 & 25.10 & 0.803 \\
    4dgf~\cite{fischer2024dynamic} & $\mathrm{B}^*$ & 27.48 & 0.852 & 27.81 & 0.865 & 25.16 & 0.789 & 27.14 & 0.858 & 26.22 & 0.843 \\
    OmniRe~\cite{chen2024omnire} & S+$\mathrm{B}^*$ & 29.15 & 0.873 & 29.54 & 0.886 & 26.10 & 0.835 & 27.23 & 0.861 & 26.68 & 0.856 \\
    \bottomrule
    \end{tabular}%
    }
\end{table*}

\section{Experiments}

\subsection{Implementation Details}
We initialize Gaussian points from both LiDAR points (with projected RGB and semantic values) and 200K random points sampled within a sphere. 
Following previous 3DGS-based methods to predict reliable semantic Gaussians, we use SegFormer~\cite{xie2021segformer} as our pre-trained segmentation model. Our approach builds upon PVG~\cite{chen2023periodic}, with all parameters configured identically to its original implementation. The learning rate of MLP starts from $1.6 \times 10^{-4}$ and decreases to $1.6 \times 10^{-6}$. For each loss term, the weighting coefficients are empirically set to $\lambda_1=0.8$, $\lambda_2=0.2$, $\lambda_3=0.01$,$\lambda_4=0.01$, $\lambda_5=0.0001$, $\lambda_6=0.1$ and $\lambda_7=0.01$. All experiments are conducted on a single NVIDIA V100s.

\subsection{Datasets}
Our experiments are conducted on two widely-used autonomous driving datasets: nuScenes~\cite{caesar2020nuscenes} and PandaSet~\cite{chen2023periodic}. The nuScenes dataset~\cite{caesar2020nuscenes} comprises 1,000 scenes captured in Boston and Singapore under diverse urban scenarios. PandaSet~\cite{chen2023periodic} is a comprehensive dataset collected in San Francisco, containing 103 sequences with synchronized LiDAR and camera data. Both datasets provide ground-truth (GT) 3D bounding boxes. To ensure fair comparisons, we also utilize a pretrained state-of-the-art 3D multi-object tracking model, MCTrack~\cite{wang2024mctrack}, to generate predicted 3D bounding boxes for methods requiring such inputs. Additionally, we conducted experiments on the Waymo dataset~\cite{sun2020scalability}; detailed results and analyses are available in the supplementary material.

\subsection{Results and Comparisons}
\noindent\textbf{Results on nuScenes~\cite{caesar2020nuscenes}.}
We comprehensively evaluate our method against previous state-of-the-art approaches on the nuScenes~\cite{caesar2020nuscenes} dataset, reporting quantitative results in Table~\ref{tab:comparison}. Our approach achieves superior performance among methods that do not rely on ground-truth (GT) bounding boxes, surpassing recent semantic-based methods (e.g., PVG~\cite{chen2023periodic}, EmerNeRF~\cite{yang2023emernerf}) as well as bounding-box-based approaches utilizing predicted boxes (e.g., streetGS~\cite{yan2024street}, OmniRe~\cite{chen2024omnire}, 4dgf~\cite{fischer2024dynamic}). Specifically, for full-image reconstruction, our method achieves 26.92 PSNR and 0.848 SSIM, significantly outperforming PVG~\cite{chen2023periodic} by 0.69 PSNR and 0.014 SSIM. In non-sky regions, our gains become even more pronounced, reaching 27.61 PSNR and 0.861 SSIM, clearly exceeding PVG by 0.89 PSNR and 0.020 SSIM. Such improvements highlight our method's effectiveness in accurately reconstructing detailed urban structures.

Remarkably, our approach not only outperforms all methods relying on predicted bounding boxes, but also surpasses the streetGS~\cite{yan2024street} method even when it leverages GT bounding boxes. This result demonstrates the strong robustness and effectiveness of our method, as we achieve superior accuracy without explicit bounding-box supervision. Additionally, methods that rely heavily on bounding boxes experience significant performance degradation when switching from GT to predicted bounding boxes; notably, the 4dgf~\cite{fischer2024dynamic} method struggles significantly under predicted bounding-box inputs, failing to converge and exhibiting very low performance.

In dynamic object reconstruction, our method further demonstrates clear advantages. For human instances, we achieve 24.92 PSNR and 0.767 SSIM, substantially outperforming PVG by 0.94 PSNR and 0.024 SSIM. For vehicle instances, we achieve 26.01 PSNR and 0.838 SSIM, representing a significant gain of 1.28 PSNR and 0.023 SSIM compared to PVG. These improvements underscore our model's strength in effectively reconstructing dynamic content in challenging urban environments, without relying on external bounding-box supervision.

We present qualitative comparisons in Fig.~\ref{first}. For dynamic objects like vehicles and pedestrians, our method shows significant improvements over PVG~\cite{chen2023periodic} and DeformGS~\cite{yang2024deformable}, with notably reduced motion blur. Meanwhile, our approach also demonstrates better reconstruction quality for static scene elements such as roads, preserving more detailed textures and geometric structures. These visual improvements align well with our quantitative results, where we achieve consistently higher scores across both dynamic and static regions. 
The qualitative results in Fig.~\ref{fig:comparison} further support these findings. 
In consecutive frame reconstruction, DeformGS~\cite{yang2024deformable} fails to handle dynamic objects, producing severe artifacts. While PVG~\cite{chen2023periodic} captures the overall shape of moving vehicles, it suffers from noticeable blurring artifacts, particularly in the lower parts of the vehicles. In contrast, our method achieves clearer and more consistent reconstruction across all dynamic objects.


\begin{table}[!t]
    \fontsize{9}{10}\selectfont
    \centering
    \renewcommand\arraystretch{1.2}
\caption{Quantitative comparison with state-of-the-art methods on the PandaSet~\cite{xiao2021pandaset} dataset. We report image reconstruction and novel view synthesis metrics. * indicates using ground-truth data}
\label{tab:comparison_pandaset}
\resizebox{\linewidth}{!}{
\begin{tabular}{l|cc|cc}
\toprule
\multirow{2}{*}{Method} & \multicolumn{2}{c|}{Image Reconstruction} & \multicolumn{2}{c}{Novel View Synthesis} \\
& PSNR$\uparrow$ & SSIM$\uparrow$ & PSNR$\uparrow$ & SSIM$\uparrow$ \\
\midrule
3DGS~\cite{kerbl20233d} & 23.67 & 0.743 & 22.14 & 0.713 \\
EmerNeRF~\cite{yang2023emernerf} & 26.45 & 0.812 & 24.89 & 0.765 \\
PVG~\cite{chen2023periodic} & 27.15 & 0.836 & 25.92 & 0.798 \\
OmniRe~\cite{chen2024omnire} & 26.94 & 0.840 & 25.75 & 0.804 \\
\midrule
\textbf{UrbanGS} (Ours) & \textbf{28.03} & \textbf{0.858} & \textbf{26.76} & \textbf{0.821} \\
\midrule 
$\mathrm{OmniRe}^*$~\cite{chen2024omnire} & 29.02 & 0.882 & 27.73 & 0.855 \\
\bottomrule

\end{tabular}
}
\end{table}

\noindent\textbf{Results on PandaSet~\cite{xiao2021pandaset}.}
We further evaluate our method on the PandaSet dataset~\cite{xiao2021pandaset}, comparing it against recent state-of-the-art methods in both image reconstruction and novel view synthesis tasks (Table~\ref{tab:comparison_pandaset}). Our approach consistently outperforms previous methods that rely on predicted bounding boxes or do not use bounding boxes at all, including OmniRe~\cite{chen2024omnire}, PVG~\cite{chen2023periodic}, EmerNeRF~\cite{yang2023emernerf}, and 3DGS~\cite{kerbl20233d}. Although methods utilizing ground-truth bounding boxes achieve higher performance, our method demonstrates superior robustness and effectiveness under realistic conditions (i.e., without ground-truth bounding boxes). These results highlight the effectiveness of our proposed approach in accurately reconstructing urban scenes. For additional qualitative results, please refer to the Supplementary.

\subsection{Ablation Study} 
We conduct an ablation study to evaluate the contribution of each module in our method on the PandaSet~\cite{chen2023periodic}. The results are summarized in Table~\ref{tab:ablation_road}.

\noindent\textbf{Effect of Static Invariance.} The absence of the static invariance module results in a notable decline in performance, with PSNR and SSIM dropping to 26.56 and 0.814, respectively. This underscores the importance of incorporating static priors to improve the reconstruction of static regions, thereby ensuring more robust and accurate results.

\noindent\textbf{Effect of Road Consistency.} Removing the road consistency module causes a performance drop. This demonstrates that enforcing road consistency is essential for preserving the geometric and textural integrity of roads, which significantly enhances the overall reconstruction quality.

\noindent\textbf{Effect of 4D Representation.} Excluding the 4D representation leads to the most significant performance degradation, with PSNR reducing to 26.43 and SSIM dropping to 0.810. This highlights the critical contribution of the 4D representation in modeling potentially dynamic objects and handling temporal variations, which are essential for reconstructing complex urban scenes.

\noindent\textbf{Visualization of Static Regularization.} Figure~\ref{fig:comparison} presents a visual comparison of the effect of static regularization on Waymo~\cite{sun2020scalability}. When the Static Invariance and Road Consistency modules are removed, the reconstructed ground becomes significantly more blurred. In contrast, incorporating these modules results in much clearer and more visually pleasing reconstructions, demonstrating their effectiveness.

\begin{table}[!t]
    \fontsize{9}{10}\selectfont
    \centering
    \renewcommand\arraystretch{1.1}
\caption{Ablation study on each module. We evaluate the effect of each module on PandaSet~\cite{chen2023periodic}.}
\label{tab:ablation_road}
\resizebox{\linewidth}{!}{
\begin{tabular}{l|cc}
\toprule
Method & PSNR$\uparrow$ & SSIM$\uparrow$ \\
\midrule
Baseline (w/o static invariance) & 26.56 & 0.814 \\
Baseline (w/o road consistency) & 26.60 & 0.816 \\
Baseline (w/o 4D Representation) & 26.43 & 0.810 \\
UrbanGS(Ours) & \textbf{26.76} & \textbf{0.821} \\
\bottomrule
\end{tabular}
}
\end{table}

\begin{figure}[!t]
    \includegraphics[width=\linewidth]{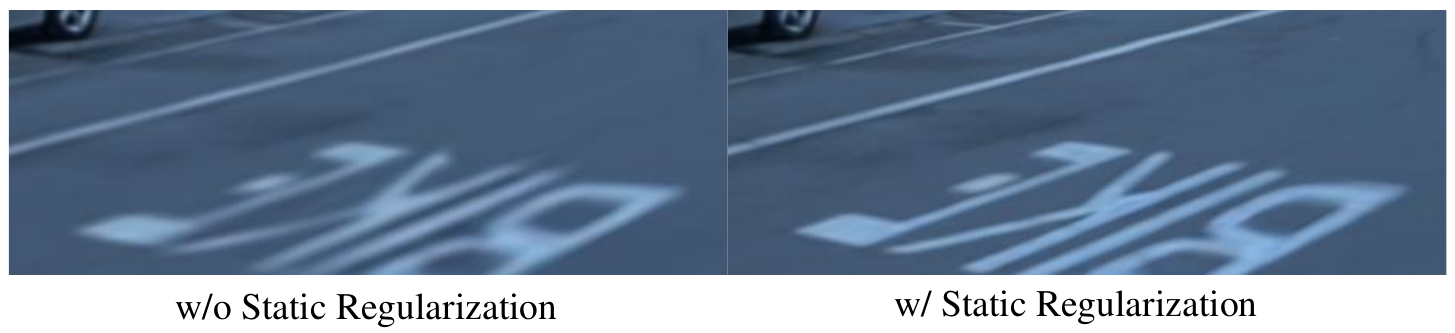}
    \caption{Ablation study on the effectiveness of static regularization. Results without static regularization (left) are blurry, while adding it (right) produces sharper details.}
    \label{fig:comparison}
\end{figure}

\section{Conclusions}
In conclusion, UrbanGS provides a novel semantic-guided decomposition strategy for reconstructing urban scenes. By leveraging 2D semantic information, our approach effectively separates static elements from potentially dynamic components. We introduced specialized processing methods for different elements:  A static Invariance module to leverage static priors to enhance the reconstruction of stationary regions and KNN-based consistency regularization for the ground surface. And a 4D Gaussian Splatting representation for potentially dynamic objects. Both qualitative and quantitative results demonstrated that UrbanGS improves rendering quality across various scene components.
\clearpage
{
    \small
    \bibliographystyle{ieeenat_fullname}
    \bibliography{main}

\begin{thebibliography}{48}
\providecommand{\natexlab}[1]{#1}
\providecommand{\url}[1]{\texttt{#1}}
\expandafter\ifx\csname urlstyle\endcsname\relax
  \providecommand{\doi}[1]{doi: #1}\else
  \providecommand{\doi}{doi: \begingroup \urlstyle{rm}\Url}\fi

\bibitem[Barron et~al.(2021)Barron, Mildenhall, Tancik, Hedman, Martin-Brualla, and Srinivasan]{barron2021mip}
Jonathan~T Barron, Ben Mildenhall, Matthew Tancik, Peter Hedman, Ricardo Martin-Brualla, and Pratul~P Srinivasan.
\newblock Mip-nerf: A multiscale representation for anti-aliasing neural radiance fields.
\newblock In \emph{Proceedings of the IEEE/CVF international conference on computer vision}, pages 5855--5864, 2021.

\bibitem[Barron et~al.(2022)Barron, Mildenhall, Verbin, Srinivasan, and Hedman]{barron2022mip}
Jonathan~T Barron, Ben Mildenhall, Dor Verbin, Pratul~P Srinivasan, and Peter Hedman.
\newblock Mip-nerf 360: Unbounded anti-aliased neural radiance fields.
\newblock In \emph{Proceedings of the IEEE/CVF conference on computer vision and pattern recognition}, pages 5470--5479, 2022.

\bibitem[Barron et~al.(2023)Barron, Mildenhall, Verbin, Srinivasan, and Hedman]{barron2023zip}
Jonathan~T Barron, Ben Mildenhall, Dor Verbin, Pratul~P Srinivasan, and Peter Hedman.
\newblock Zip-nerf: Anti-aliased grid-based neural radiance fields.
\newblock In \emph{Proceedings of the IEEE/CVF International Conference on Computer Vision}, pages 19697--19705, 2023.

\bibitem[Caesar et~al.(2020)Caesar, Bankiti, Lang, Vora, Liong, Xu, Krishnan, Pan, Baldan, and Beijbom]{caesar2020nuscenes}
Holger Caesar, Varun Bankiti, Alex~H Lang, Sourabh Vora, Venice~Erin Liong, Qiang Xu, Anush Krishnan, Yu Pan, Giancarlo Baldan, and Oscar Beijbom.
\newblock nuscenes: A multimodal dataset for autonomous driving.
\newblock In \emph{Proceedings of the IEEE/CVF conference on computer vision and pattern recognition}, pages 11621--11631, 2020.

\bibitem[Cai et~al.(2022)Cai, Feng, Feng, Wang, and Zhang]{cai2022neural}
Hongrui Cai, Wanquan Feng, Xuetao Feng, Yan Wang, and Juyong Zhang.
\newblock Neural surface reconstruction of dynamic scenes with monocular rgb-d camera.
\newblock \emph{Advances in Neural Information Processing Systems}, 35:\penalty0 967--981, 2022.

\bibitem[Chen et~al.(2022)Chen, Xu, Geiger, Yu, and Su]{chen2022tensorf}
Anpei Chen, Zexiang Xu, Andreas Geiger, Jingyi Yu, and Hao Su.
\newblock Tensorf: Tensorial radiance fields.
\newblock In \emph{European conference on computer vision}, pages 333--350. Springer, 2022.

\bibitem[Chen et~al.(2023)Chen, Gu, Jiang, Zhu, and Zhang]{chen2023periodic}
Yurui Chen, Chun Gu, Junzhe Jiang, Xiatian Zhu, and Li Zhang.
\newblock Periodic vibration gaussian: Dynamic urban scene reconstruction and real-time rendering.
\newblock \emph{arXiv preprint arXiv:2311.18561}, 2023.

\bibitem[Chen et~al.(2024)Chen, Yang, Huang, de~Lutio, Esturo, Ivanovic, Litany, Gojcic, Fidler, Pavone, et~al.]{chen2024omnire}
Ziyu Chen, Jiawei Yang, Jiahui Huang, Riccardo de Lutio, Janick~Martinez Esturo, Boris Ivanovic, Or Litany, Zan Gojcic, Sanja Fidler, Marco Pavone, et~al.
\newblock Omnire: Omni urban scene reconstruction.
\newblock \emph{arXiv preprint arXiv:2408.16760}, 2024.

\bibitem[Cheng et~al.(2024)Cheng, Long, Yang, Yao, Yin, Ma, Wang, and Chen]{cheng2024gaussianpro}
Kai Cheng, Xiaoxiao Long, Kaizhi Yang, Yao Yao, Wei Yin, Yuexin Ma, Wenping Wang, and Xuejin Chen.
\newblock Gaussianpro: 3d gaussian splatting with progressive propagation.
\newblock In \emph{Forty-first International Conference on Machine Learning}, 2024.

\bibitem[Fischer et~al.(2024)Fischer, Kulhanek, Bulo, Porzi, Pollefeys, and Kontschieder]{fischer2024dynamic}
Tobias Fischer, Jonas Kulhanek, Samuel~Rota Bulo, Lorenzo Porzi, Marc Pollefeys, and Peter Kontschieder.
\newblock Dynamic 3d gaussian fields for urban areas.
\newblock \emph{arXiv preprint arXiv:2406.03175}, 2024.

\bibitem[Fridovich-Keil et~al.(2022)Fridovich-Keil, Yu, Tancik, Chen, Recht, and Kanazawa]{fridovich2022plenoxels}
Sara Fridovich-Keil, Alex Yu, Matthew Tancik, Qinhong Chen, Benjamin Recht, and Angjoo Kanazawa.
\newblock Plenoxels: Radiance fields without neural networks.
\newblock In \emph{Proceedings of the IEEE/CVF conference on computer vision and pattern recognition}, pages 5501--5510, 2022.

\bibitem[Garbin et~al.(2021)Garbin, Kowalski, Johnson, Shotton, and Valentin]{garbin2021fastnerf}
Stephan~J Garbin, Marek Kowalski, Matthew Johnson, Jamie Shotton, and Julien Valentin.
\newblock Fastnerf: High-fidelity neural rendering at 200fps.
\newblock In \emph{Proceedings of the IEEE/CVF international conference on computer vision}, pages 14346--14355, 2021.

\bibitem[Guo et~al.(2024)Guo, Ma, Fan, Liu, and Li]{guo2024semantic}
Jun Guo, Xiaojian Ma, Yue Fan, Huaping Liu, and Qing Li.
\newblock Semantic gaussians: Open-vocabulary scene understanding with 3d gaussian splatting.
\newblock \emph{arXiv preprint arXiv:2403.15624}, 2024.

\bibitem[Guo et~al.(2022)Guo, Kang, Bao, He, and Zhang]{guo2022nerfren}
Yuan-Chen Guo, Di Kang, Linchao Bao, Yu He, and Song-Hai Zhang.
\newblock Nerfren: Neural radiance fields with reflections.
\newblock In \emph{Proceedings of the IEEE/CVF Conference on Computer Vision and Pattern Recognition}, pages 18409--18418, 2022.

\bibitem[Huang et~al.(2024)Huang, Wei, Zheng, An, Lu, Zhan, Tomizuka, Keutzer, and Zhang]{huang2024s3gaussian}
Nan Huang, Xiaobao Wei, Wenzhao Zheng, Pengju An, Ming Lu, Wei Zhan, Masayoshi Tomizuka, Kurt Keutzer, and Shanghang Zhang.
\newblock S3gaussian: Self-supervised street gaussians for autonomous driving.
\newblock \emph{arXiv preprint arXiv:2405.20323}, 2024.

\bibitem[Jiang et~al.(2024)Jiang, Tu, Liu, Gao, Long, Wang, and Ma]{jiang2024gaussianshader}
Yingwenqi Jiang, Jiadong Tu, Yuan Liu, Xifeng Gao, Xiaoxiao Long, Wenping Wang, and Yuexin Ma.
\newblock Gaussianshader: 3d gaussian splatting with shading functions for reflective surfaces.
\newblock In \emph{Proceedings of the IEEE/CVF Conference on Computer Vision and Pattern Recognition}, pages 5322--5332, 2024.

\bibitem[Kerbl et~al.(2023)Kerbl, Kopanas, Leimk{\"u}hler, and Drettakis]{kerbl20233d}
Bernhard Kerbl, Georgios Kopanas, Thomas Leimk{\"u}hler, and George Drettakis.
\newblock 3d gaussian splatting for real-time radiance field rendering.
\newblock \emph{ACM Trans. Graph.}, 42\penalty0 (4):\penalty0 139--1, 2023.

\bibitem[Kopanas et~al.(2022)Kopanas, Leimk{\"u}hler, Rainer, Jambon, and Drettakis]{kopanas2022neural}
Georgios Kopanas, Thomas Leimk{\"u}hler, Gilles Rainer, Cl{\'e}ment Jambon, and George Drettakis.
\newblock Neural point catacaustics for novel-view synthesis of reflections.
\newblock \emph{ACM Transactions on Graphics (TOG)}, 41\penalty0 (6):\penalty0 1--15, 2022.

\bibitem[Li et~al.(2022)Li, Slavcheva, Zollhoefer, Green, Lassner, Kim, Schmidt, Lovegrove, Goesele, Newcombe, et~al.]{li2022neural}
Tianye Li, Mira Slavcheva, Michael Zollhoefer, Simon Green, Christoph Lassner, Changil Kim, Tanner Schmidt, Steven Lovegrove, Michael Goesele, Richard Newcombe, et~al.
\newblock Neural 3d video synthesis from multi-view video.
\newblock In \emph{Proceedings of the IEEE/CVF Conference on Computer Vision and Pattern Recognition}, pages 5521--5531, 2022.

\bibitem[Li et~al.(2021)Li, Niklaus, Snavely, and Wang]{li2021neural}
Zhengqi Li, Simon Niklaus, Noah Snavely, and Oliver Wang.
\newblock Neural scene flow fields for space-time view synthesis of dynamic scenes.
\newblock In \emph{Proceedings of the IEEE/CVF Conference on Computer Vision and Pattern Recognition}, pages 6498--6508, 2021.

\bibitem[Liu et~al.(2024)Liu, Hu, Yang, Chen, Wang, Chen, Cai, Gao, and Zhao]{liu2024rip}
Junchen Liu, Wenbo Hu, Zhuo Yang, Jianteng Chen, Guoliang Wang, Xiaoxue Chen, Yantong Cai, Huan-ang Gao, and Hao Zhao.
\newblock Rip-nerf: Anti-aliasing radiance fields with ripmap-encoded platonic solids.
\newblock In \emph{ACM SIGGRAPH 2024 Conference Papers}, pages 1--11, 2024.

\bibitem[Luiten et~al.(2024)Luiten, Kopanas, Leibe, and Ramanan]{luiten2023dynamic}
Jonathon Luiten, Georgios Kopanas, Bastian Leibe, and Deva Ramanan.
\newblock Dynamic 3d gaussians: Tracking by persistent dynamic view synthesis.
\newblock In \emph{3DV}, 2024.

\bibitem[Martin-Brualla et~al.(2021)Martin-Brualla, Radwan, Sajjadi, Barron, Dosovitskiy, and Duckworth]{martin2021nerf}
Ricardo Martin-Brualla, Noha Radwan, Mehdi~SM Sajjadi, Jonathan~T Barron, Alexey Dosovitskiy, and Daniel Duckworth.
\newblock Nerf in the wild: Neural radiance fields for unconstrained photo collections.
\newblock In \emph{Proceedings of the IEEE/CVF conference on computer vision and pattern recognition}, pages 7210--7219, 2021.

\bibitem[Mildenhall et~al.(2021)Mildenhall, Srinivasan, Tancik, Barron, Ramamoorthi, and Ng]{mildenhall2021nerf}
Ben Mildenhall, Pratul~P Srinivasan, Matthew Tancik, Jonathan~T Barron, Ravi Ramamoorthi, and Ren Ng.
\newblock Nerf: Representing scenes as neural radiance fields for view synthesis.
\newblock \emph{Communications of the ACM}, 65\penalty0 (1):\penalty0 99--106, 2021.

\bibitem[M\"uller et~al.(2022)M\"uller, Evans, Schied, and Keller]{mueller2022instant}
Thomas M\"uller, Alex Evans, Christoph Schied, and Alexander Keller.
\newblock Instant neural graphics primitives with a multiresolution hash encoding.
\newblock \emph{ACM Trans. Graph.}, 41\penalty0 (4):\penalty0 102:1--102:15, 2022.

\bibitem[Park et~al.(2021{\natexlab{a}})Park, Sinha, Barron, Bouaziz, Goldman, Seitz, and Martin-Brualla]{park2021nerfies}
Keunhong Park, Utkarsh Sinha, Jonathan~T Barron, Sofien Bouaziz, Dan~B Goldman, Steven~M Seitz, and Ricardo Martin-Brualla.
\newblock Nerfies: Deformable neural radiance fields.
\newblock In \emph{Proceedings of the IEEE/CVF International Conference on Computer Vision}, pages 5865--5874, 2021{\natexlab{a}}.

\bibitem[Park et~al.(2021{\natexlab{b}})Park, Sinha, Hedman, Barron, Bouaziz, Goldman, Martin-Brualla, and Seitz]{park2021hypernerf}
Keunhong Park, Utkarsh Sinha, Peter Hedman, Jonathan~T Barron, Sofien Bouaziz, Dan~B Goldman, Ricardo Martin-Brualla, and Steven~M Seitz.
\newblock Hypernerf: A higher-dimensional representation for topologically varying neural radiance fields.
\newblock \emph{arXiv preprint arXiv:2106.13228}, 2021{\natexlab{b}}.

\bibitem[Pumarola et~al.(2021)Pumarola, Corona, Pons-Moll, and Moreno-Noguer]{pumarola2021d}
Albert Pumarola, Enric Corona, Gerard Pons-Moll, and Francesc Moreno-Noguer.
\newblock D-nerf: Neural radiance fields for dynamic scenes.
\newblock In \emph{Proceedings of the IEEE/CVF Conference on Computer Vision and Pattern Recognition}, pages 10318--10327, 2021.

\bibitem[Sun et~al.(2022)Sun, Sun, and Chen]{sun2022direct}
Cheng Sun, Min Sun, and Hwann-Tzong Chen.
\newblock Direct voxel grid optimization: Super-fast convergence for radiance fields reconstruction.
\newblock In \emph{Proceedings of the IEEE/CVF conference on computer vision and pattern recognition}, pages 5459--5469, 2022.

\bibitem[Sun et~al.(2020)Sun, Kretzschmar, Dotiwalla, Chouard, Patnaik, Tsui, Guo, Zhou, Chai, Caine, et~al.]{sun2020scalability}
Pei Sun, Henrik Kretzschmar, Xerxes Dotiwalla, Aurelien Chouard, Vijaysai Patnaik, Paul Tsui, James Guo, Yin Zhou, Yuning Chai, Benjamin Caine, et~al.
\newblock Scalability in perception for autonomous driving: Waymo open dataset.
\newblock In \emph{Proceedings of the IEEE/CVF conference on computer vision and pattern recognition}, pages 2446--2454, 2020.

\bibitem[Tancik et~al.(2022)Tancik, Casser, Yan, Pradhan, Mildenhall, Srinivasan, Barron, and Kretzschmar]{tancik2022block}
Matthew Tancik, Vincent Casser, Xinchen Yan, Sabeek Pradhan, Ben Mildenhall, Pratul~P Srinivasan, Jonathan~T Barron, and Henrik Kretzschmar.
\newblock Block-nerf: Scalable large scene neural view synthesis.
\newblock In \emph{Proceedings of the IEEE/CVF Conference on Computer Vision and Pattern Recognition}, pages 8248--8258, 2022.

\bibitem[Tonderski et~al.(2024)Tonderski, Lindstr{\"o}m, Hess, Ljungbergh, Svensson, and Petersson]{tonderski2024neurad}
Adam Tonderski, Carl Lindstr{\"o}m, Georg Hess, William Ljungbergh, Lennart Svensson, and Christoffer Petersson.
\newblock Neurad: Neural rendering for autonomous driving.
\newblock In \emph{Proceedings of the IEEE/CVF Conference on Computer Vision and Pattern Recognition}, pages 14895--14904, 2024.

\bibitem[Tretschk et~al.(2021)Tretschk, Tewari, Golyanik, Zollh{\"o}fer, Lassner, and Theobalt]{tretschk2021non}
Edgar Tretschk, Ayush Tewari, Vladislav Golyanik, Michael Zollh{\"o}fer, Christoph Lassner, and Christian Theobalt.
\newblock Non-rigid neural radiance fields: Reconstruction and novel view synthesis of a dynamic scene from monocular video.
\newblock In \emph{Proceedings of the IEEE/CVF International Conference on Computer Vision}, pages 12959--12970, 2021.

\bibitem[Turki et~al.(2022)Turki, Ramanan, and Satyanarayanan]{turki2022mega}
Haithem Turki, Deva Ramanan, and Mahadev Satyanarayanan.
\newblock Mega-nerf: Scalable construction of large-scale nerfs for virtual fly-throughs.
\newblock In \emph{Proceedings of the IEEE/CVF Conference on Computer Vision and Pattern Recognition}, pages 12922--12931, 2022.

\bibitem[Turki et~al.(2023)Turki, Zhang, Ferroni, and Ramanan]{turki2023suds}
Haithem Turki, Jason~Y Zhang, Francesco Ferroni, and Deva Ramanan.
\newblock Suds: Scalable urban dynamic scenes.
\newblock In \emph{Proceedings of the IEEE/CVF Conference on Computer Vision and Pattern Recognition}, pages 12375--12385, 2023.

\bibitem[Verbin et~al.(2022)Verbin, Hedman, Mildenhall, Zickler, Barron, and Srinivasan]{verbin2022ref}
Dor Verbin, Peter Hedman, Ben Mildenhall, Todd Zickler, Jonathan~T Barron, and Pratul~P Srinivasan.
\newblock Ref-nerf: Structured view-dependent appearance for neural radiance fields.
\newblock In \emph{2022 IEEE/CVF Conference on Computer Vision and Pattern Recognition (CVPR)}, pages 5481--5490. IEEE, 2022.

\bibitem[Wang et~al.(2024)Wang, Qi, Zhao, Zhou, Zhang, Wang, Tu, Guo, Zhao, Li, et~al.]{wang2024mctrack}
Xiyang Wang, Shouzheng Qi, Jieyou Zhao, Hangning Zhou, Siyu Zhang, Guoan Wang, Kai Tu, Songlin Guo, Jianbo Zhao, Jian Li, et~al.
\newblock Mctrack: A unified 3d multi-object tracking framework for autonomous driving.
\newblock \emph{arXiv preprint arXiv:2409.16149}, 2024.

\bibitem[Wu et~al.(2023)Wu, Liu, Luo, Zhong, Chen, Xiao, Hou, Lou, Chen, Yang, et~al.]{wu2023mars}
Zirui Wu, Tianyu Liu, Liyi Luo, Zhide Zhong, Jianteng Chen, Hongmin Xiao, Chao Hou, Haozhe Lou, Yuantao Chen, Runyi Yang, et~al.
\newblock Mars: An instance-aware, modular and realistic simulator for autonomous driving.
\newblock In \emph{CAAI International Conference on Artificial Intelligence}, pages 3--15. Springer, 2023.

\bibitem[Xian et~al.(2021)Xian, Huang, Kopf, and Kim]{xian2021space}
Wenqi Xian, Jia-Bin Huang, Johannes Kopf, and Changil Kim.
\newblock Space-time neural irradiance fields for free-viewpoint video.
\newblock In \emph{Proceedings of the IEEE/CVF conference on computer vision and pattern recognition}, pages 9421--9431, 2021.

\bibitem[Xiao et~al.(2021)Xiao, Shao, Hao, Zhang, Chai, Jiao, Li, Wu, Sun, Jiang, et~al.]{xiao2021pandaset}
Pengchuan Xiao, Zhenlei Shao, Steven Hao, Zishuo Zhang, Xiaolin Chai, Judy Jiao, Zesong Li, Jian Wu, Kai Sun, Kun Jiang, et~al.
\newblock Pandaset: Advanced sensor suite dataset for autonomous driving.
\newblock In \emph{2021 IEEE International Intelligent Transportation Systems Conference (ITSC)}, pages 3095--3101. IEEE, 2021.

\bibitem[Xie et~al.(2021)Xie, Wang, Yu, Anandkumar, Alvarez, and Luo]{xie2021segformer}
Enze Xie, Wenhai Wang, Zhiding Yu, Anima Anandkumar, Jose~M Alvarez, and Ping Luo.
\newblock Segformer: Simple and efficient design for semantic segmentation with transformers.
\newblock \emph{Advances in neural information processing systems}, 34:\penalty0 12077--12090, 2021.

\bibitem[Xie et~al.(2023)Xie, Zhang, Li, Zhang, and Zhang]{xie2023s}
Ziyang Xie, Junge Zhang, Wenye Li, Feihu Zhang, and Li Zhang.
\newblock S-nerf: Neural radiance fields for street views.
\newblock \emph{arXiv preprint arXiv:2303.00749}, 2023.

\bibitem[Yan et~al.(2024)Yan, Lin, Zhou, Wang, Sun, Zhan, Lang, Zhou, and Peng]{yan2024street}
Yunzhi Yan, Haotong Lin, Chenxu Zhou, Weijie Wang, Haiyang Sun, Kun Zhan, Xianpeng Lang, Xiaowei Zhou, and Sida Peng.
\newblock Street gaussians: Modeling dynamic urban scenes with gaussian splatting.
\newblock In \emph{ECCV}, 2024.

\bibitem[Yang et~al.(2023{\natexlab{a}})Yang, Ivanovic, Litany, Weng, Kim, Li, Che, Xu, Fidler, Pavone, et~al.]{yang2023emernerf}
Jiawei Yang, Boris Ivanovic, Or Litany, Xinshuo Weng, Seung~Wook Kim, Boyi Li, Tong Che, Danfei Xu, Sanja Fidler, Marco Pavone, et~al.
\newblock Emernerf: Emergent spatial-temporal scene decomposition via self-supervision.
\newblock \emph{arXiv preprint arXiv:2311.02077}, 2023{\natexlab{a}}.

\bibitem[Yang et~al.(2023{\natexlab{b}})Yang, Chen, Wang, Manivasagam, Ma, Yang, and Urtasun]{yang2023unisim}
Ze Yang, Yun Chen, Jingkang Wang, Sivabalan Manivasagam, Wei-Chiu Ma, Anqi~Joyce Yang, and Raquel Urtasun.
\newblock Unisim: A neural closed-loop sensor simulator.
\newblock In \emph{Proceedings of the IEEE/CVF Conference on Computer Vision and Pattern Recognition}, pages 1389--1399, 2023{\natexlab{b}}.

\bibitem[Yang et~al.(2024)Yang, Gao, Zhou, Jiao, Zhang, and Jin]{yang2024deformable}
Ziyi Yang, Xinyu Gao, Wen Zhou, Shaohui Jiao, Yuqing Zhang, and Xiaogang Jin.
\newblock Deformable 3d gaussians for high-fidelity monocular dynamic scene reconstruction.
\newblock In \emph{Proceedings of the IEEE/CVF Conference on Computer Vision and Pattern Recognition}, pages 20331--20341, 2024.

\bibitem[Zhou et~al.(2024{\natexlab{a}})Zhou, Chang, Jiang, Fan, Zhu, Xu, Chari, You, Wang, and Kadambi]{zhou2024feature}
Shijie Zhou, Haoran Chang, Sicheng Jiang, Zhiwen Fan, Zehao Zhu, Dejia Xu, Pradyumna Chari, Suya You, Zhangyang Wang, and Achuta Kadambi.
\newblock Feature 3dgs: Supercharging 3d gaussian splatting to enable distilled feature fields.
\newblock In \emph{Proceedings of the IEEE/CVF Conference on Computer Vision and Pattern Recognition}, pages 21676--21685, 2024{\natexlab{a}}.

\bibitem[Zhou et~al.(2024{\natexlab{b}})Zhou, Lin, Shan, Wang, Sun, and Yang]{zhou2024drivinggaussian}
Xiaoyu Zhou, Zhiwei Lin, Xiaojun Shan, Yongtao Wang, Deqing Sun, and Ming-Hsuan Yang.
\newblock Drivinggaussian: Composite gaussian splatting for surrounding dynamic autonomous driving scenes.
\newblock In \emph{Proceedings of the IEEE/CVF Conference on Computer Vision and Pattern Recognition}, pages 21634--21643, 2024{\natexlab{b}}.

\end{thebibliography}
}


\end{document}